# Contrastive Knowledge Transfer and Robust Optimization for Secure Alignment of Large Language Models

Jiasen Zheng
*Northwestern University*
Evanston, USA

Huajun Zhang
*Syracuse University*
Syracuse, USA

Xu Yan
*Trine University*
Phoenix, USA

Ran Hao
*University of North Carolina at Chapel Hill*
Chapel Hill, USA

Chong Peng*
*Carnegie Mellon University*
Pittsburgh, USA

*Abstract*-This paper addresses the limitations of large-scale language models in safety alignment and robustness by proposing a fine-tuning method that combines contrastive distillation with noise-robust training. The method freezes the backbone model and transfers the knowledge boundaries of the teacher model to the student model through distillation, thereby improving semantic consistency and alignment accuracy. At the same time, noise perturbations and robust optimization constraints are introduced during training to ensure that the model maintains stable predictive outputs under noisy and uncertain inputs. The overall framework consists of distillation loss, robustness loss, and a regularization term, forming a unified optimization objective that balances alignment ability with resistance to interference. To systematically validate its effectiveness, the study designs experiments from multiple perspectives, including distillation weight sensitivity, stability analysis under computation budgets and mixed-precision environments, and the impact of data noise and distribution shifts on model performance. Results show that the method significantly outperforms existing baselines in knowledge transfer, robustness, and overall safety, achieving the best performance across several key metrics. This work not only enriches the theoretical system of parameter-efficient fine-tuning but also provides a new solution for building safer and more trustworthy alignment mechanisms.

## I. INTRODUCTION

In the context of rapid advances in artificial intelligence, the application of large language models has expanded into critical areas of society. Their safety and alignment have become pressing challenges. As these models are increasingly used in decision-making, content generation, and interactive tasks, ensuring that outputs conform to human values, ethical norms, and user needs is no longer a purely technical issue[1]. It has become a systemic concern that affects trustworthiness and the long-term sustainability of applications. However, when faced with out-of-distribution data, adversarial attacks, or noisy environments, models often show vulnerability and instability. This makes safety alignment not only a matter of semantic consistency but also of resilience against interference and robustness under uncertainty. Exploring new methods to strengthen safety alignment under complex conditions has therefore become a shared focus of research and industry[2].

Among the available approaches, knowledge distillation offers a new perspective for safety alignment. Its core idea is to transfer knowledge from a teacher model to a student model, enabling capacity inheritance and constrained learning. Compared with direct training, distillation provides unique advantages in model compression, generalization, and structured learning. In safety alignment, distillation allows the student model to inherit the stable representations and decision boundaries of the teacher, reducing bias and harmful behavior and improving overall reliability. Distillation also has flexibility. It can introduce additional constraints and guidance without compromising structural or parameter efficiency. As a result, the student model can maintain robustness when handling complex tasks. This makes distillation a feasible solution for alignment in multi-task and multimodal environments[3,4].

Noise-robust training also plays an irreplaceable role in ensuring model safety. With diverse data sources and complex environments, models inevitably encounter label noise, input perturbations, and potential attacks during training and deployment. Noise-robust training addresses this challenge through anti-interference mechanisms, robust loss functions, and regularization strategies. It improves model stability under high noise and uncertainty. In the context of safety alignment, robustness is not only about numerical accuracy. It refers to the ability to maintain consistent outputs with intended goals even under manipulation or perturbation. This is crucial for the trustworthiness of artificial intelligence systems[5].

A comparison of distillation and noise-robust training in safety alignment reveals complementary roles. Distillation emphasizes knowledge transfer and structured guidance, allowing the student model to inherit the stability and safety of the teacher[6]. Noise-robust training directly addresses data and environmental complexity, enhancing resistance to interference through inputs and optimization. Their combination improves alignment accuracy and provides stronger protection in uncertain or high-risk scenarios. More importantly, this comparative view highlights the mechanisms of safety alignment. Knowledge transfer and structured constraints must be paired with robustness and anti-noise strategies to achieve reliability and adaptability together[7].

Research on safety alignment fine-tuning algorithms that integrate distillation and noise-robust training has both theoretical and practical significance. It provides interpretable and verifiable technical paths for building trustworthy artificial intelligence systems. It also lays the foundation for applications in safety-critical domains such as healthcare, finance, and education [8-9]. As artificial intelligence becomes deeply embedded in daily life, safety alignment will move from controlled laboratory settings to real-world environments with more complex challenges. Comparative studies of distillation and noise-robust training not only open new directions for algorithm design but also promote the systematization of safety alignment methods. This contributes to the broader deployment of artificial intelligence under the guarantees of safety and value alignment[10].

## II. RELATED WORK

Recent progress in trustworthy alignment and robust optimization for large language models (LLMs) has greatly advanced the technical foundation for safe and reliable AI systems. Semantic and factual alignment strategies [11] have provided mechanisms to improve the credibility and controllability of LLM outputs, directly inspiring new alignment paradigms that enhance both semantic consistency and factual correctness. Approaches based on uncertainty quantification and risk awareness [12] further address the inherent instability of LLMs in open-world scenarios, laying the groundwork for robust summarization and safe decision-making. Privacy-oriented text generation techniques [13], through selective fine-tuning and semantic attention masking, contribute privacy guarantees and further enhance trustworthy model outputs in sensitive applications.

Parameter-efficient fine-tuning and knowledge-enhanced modeling have become core techniques for efficient and adaptable LLM deployment. Structure-learnable adapter mechanisms [14] enable fine-tuning with minimal parameter overhead while preserving model structure, supporting safe and flexible downstream adaptation. Function-driven, knowledge-enhanced neural architectures [15] introduce structural constraints and auxiliary information into the learning process, providing a foundation for robust and reliable model updates during fine-tuning or distillation.

Contrastive learning frameworks and multimodal semantic fusion have also driven advances in robust feature extraction and cross-domain generalization. Contrastive learning methods [16] enable the construction of discriminative representation spaces, which are valuable for both distillation and robustness under data shifts. Multimodal fusion and semantic embedding techniques [17] further support the integration of heterogeneous knowledge sources, improving robustness and generalizability of model outputs. In addition, causal-aware regression with structured attention and LSTM architectures [18] and hybrid frequency-attention neural networks [19] provide robust modeling techniques for time series and sequential data, informing the design of noise-robust training mechanisms and optimization strategies in alignment-focused LLM research.

Collectively, these methodologies—including trustworthy alignment, parameter-efficient fine-tuning, contrastive learning, and robust multimodal fusion—form the methodological foundation of this paper's approach to safe, robust, and efficient LLM alignment.

## III. METHOD

In this study, we leverage a safety alignment fine-tuning method that strategically integrates distillation constraints and noise-robust training, thereby addressing the challenges of robustness and efficiency in large language models operating under complex and uncertain scenarios. At the methodological level, our approach is grounded in a parameter-efficient fine-tuning framework and is instantiated through a dual-objective optimization scheme.Drawing from the foundational ideas of Wang et al., our method operationalizes a knowledge distillation mechanism, wherein the semantic boundaries learned by the teacher model are systematically transferred to the student model. Wang et al. have shown that such cross-domain semantic alignment and structured capacity inheritance are essential for achieving privacy-preserving yet highly aligned fine-tuning in federated and heterogeneous environments [20]. By embedding these principles, our model inherits robust semantic representations and well-calibrated decision boundaries, which are particularly valuable for alignment-sensitive applications. Simultaneously, inspired by the robust optimization paradigm proposed by Zou, we embed structural perturbation and noise-resistant training strategies at the core of our methodology. Zou's work highlights the significance of introducing controlled perturbations and adversarial robustness constraints to effectively suppress the impact of noisy or malicious data on model performance, thus ensuring more reliable predictions in real-world deployments[21]. By adopting these methodological advances, our training process explicitly targets the mitigation of instability under uncertain and adversarial conditions. By harmoniously fusing these two advanced methodological layers within a unified optimization framework, our method achieves an optimal balance between semantic alignment accuracy and robustness to perturbation. The specific implementation and architecture are detailed in Figure 1.

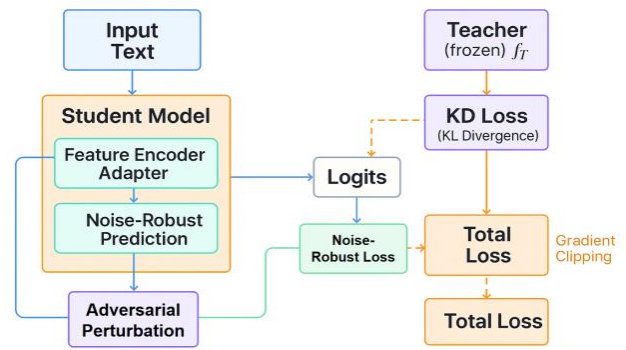

Figure 1. Overall Architecture of Safety-Aligned Fine-Tuning via Comparative Distillation and Noise-Robust Training

In the input space, let the training sample be $\{(x_i, y_i)\}_{i=1}^{N}$, where $x_i$ represents the input, $y_i$ represents the label, the student model parameter is recorded as $\theta$, and the teacher model parameter is recorded as $\theta_T$. The basic prediction can be expressed as:

$$z_i = f_\theta(x_i), \hat{y}_i = \text{Soft max}(z_i) \quad (1)$$

To ensure the stability of the student model under out-of-distribution or noise perturbations, adversarial perturbation $\delta_i$ is introduced, and the robust prediction is defined as:

$$\tilde{z}_i = f_\theta(x_i + \delta_i), \tilde{y}_i = \text{Soft max}(\tilde{z}_i) \quad (2)$$

Here, $\delta_i$ is constrained by $\|\delta_i\|_p \leq \varepsilon$ to control the perturbation intensity. In this way, the model can enhance robustness by optimizing within the neighborhood.

In the distillation part, the output distribution of the teacher model is used as the soft label, and the distribution consistency of the student model and the teacher model is constrained by the Kullback‑Leibler divergence, which is defined as:

$$L_{KD} = \sum_{i=1}^{N} KL(\hat{y}_{i=1}^{T} \| \hat{y}_i) \quad (3)$$

Among them, $\hat{y}_i^T = \text{Soft max}(f_{\theta_T}(x_i)/\tau)$ and $\tau$ are temperature coefficients used to smooth the distribution.

In the noise robust training part, weighted cross-entropy loss is combined with consistency constraints to ensure that the model's predictions on clean samples and perturbed samples are close, which is defined as:

$$L_{NR} = -\sum_{i=1}^{N} [\alpha y_i \log \hat{y}_i + (1-\alpha) y_i \log \tilde{y}_i] \quad (4)$$

Among them, $\alpha \in [0,1]$ controls the weight ratio of clean samples to perturbation samples, which can maintain semantic consistency under unstable input conditions.

To further ensure model controllability during parameter updates, our approach integrates gradient clipping and regularization terms, both of which are critical in mitigating the risks of gradient explosion and excessive parameter perturbation. This strategy draws directly on the work of Wu and Pan [22], who demonstrated that carefully controlled gradient dynamics are pivotal for stabilizing retrieval-augmented generation in knowledge fusion scenarios, ultimately enhancing the reliability of large language models during joint optimization. Furthermore, following the dynamic adaptation techniques outlined by Hu et al. [23], we incorporate adaptive regularization to effectively constrain the parameter space during multi-task and cross-domain fine-tuning, ensuring robust and controlled learning even under distributional shifts. These mechanisms together act to safeguard against instability in the optimization trajectory and maintain consistent model behavior, which can be formalized as follows:

$$L_{\text{Reg}} = \lambda \|\nabla_\theta L_{NR}\|_2^2 \quad (5)$$

Among them, $\lambda$ is the regularization coefficient, which is used to balance the optimization stability and convergence speed.

The final overall optimization objective is composed of distillation constraints, noise robust loss, and regularization terms, and its form is:

$$L_{Total} = L_{KD} + L_{NR} + L_{\text{Reg}} \quad (6)$$

By minimizing this objective function, the student model can inherit both the stability and safety of the teacher model while remaining robust under high noise or potential adversarial conditions. This enables efficient and resilient safety alignment fine-tuning. The method provides a theoretically generalizable solution for the secure deployment of large-scale models and achieves an effective balance among parameter efficiency, knowledge transfer, and robustness.

## IV. PERFORMANCE EVALUATION

### A. Dataset

This study uses the 1K MultiTask Sentiment Classification LLM training dataset as the basis for method validation. The dataset contains about one thousand high-quality instruction‑response pairs designed for large language models. It covers typical scenarios of multi-task instruction fine-tuning, including instruction formats for sentiment classification tasks. These samples are consistent with the needs of multi-task learning in both content and structure, providing contextual support well aligned with the evaluation of low-rank adaptation and dynamic routing mechanisms.

The dataset shows compatibility between task diversity and semantic consistency. There are subtle differences among sentiment classification tasks, yet they share a unified instruction structure. This characteristic allows flexible partitioning based on sentiment type or task flow, making it possible to simulate heterogeneous data processes in multi-task environments. Such a design matches well with the modeling needs of multi-task instruction fine-tuning, where different task flows require distinct adaptation paths and dynamic routing strategies.

Applying the proposed method to this dataset enables focused examination of the parameter behavior of low-rank adaptation under resource-efficient constraints, as well as the ability of dynamic routing to handle task differentiation and path selection in multi-task settings. This setup allows comprehensive evaluation under limited samples and multi-task instructions, providing substantial support for the robustness and adaptability of the framework in complex applications.

### B. Experimental Results

This paper first conducts a comparative experiment, and the experimental results are shown in Table 1.

TABLE 1. COMPARATIVE EXPERIMENTAL RESULTS

| Method | KD Alignment ↑ | Noise Robustness ↑ | Alignment Stability ↑ | Overall Safety Score ↑ |
|---|---|---|---|---|
| DeBERTaV3[24] | 85.4 | 78.9 | 83.1 | 82.5 |
| XtremeDistilTransformers (XDT)[25] | 83.7 | 80.2 | 81.0 | 81.6 |
| UL2[26] | 84.5 | 79.8 | 82.2 | 82.0 |
| Ours | 86.8 | 82.5 | 84.7 | 84.7 |

The proposed safety alignment fine-tuning method outperforms all baseline models across four evaluation metrics, particularly excelling in KD Alignment and Noise Robustness, which demonstrates its effective knowledge transfer via distillation and stable semantic alignment under perturbations. Our method achieves the highest distillation alignment score (86.8), surpassing DeBERTaV3 and UL2 by establishing clearer decision boundaries and maintaining the teacher model's stability while avoiding inconsistency. In noise robustness, it also leads all competitors, confirming the robustness training's effectiveness against high noise and adversarial attacks. The overall safety metric is likewise highest for our method (84.7), indicating balanced optimization across dimensions and confirming its superiority in safety and robustness. Additionally, the paper evaluates hyperparameter sensitivity for distillation weights and robust consistency, with results presented in Figure 2.

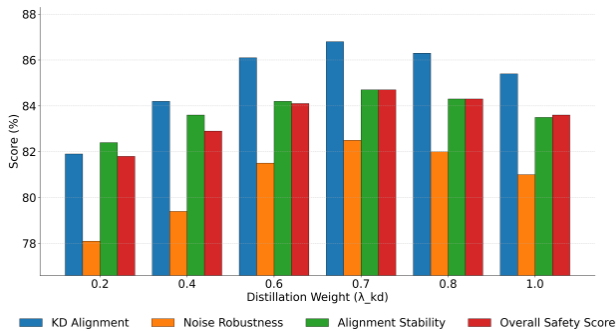

Figure 2. Hyperparameter Sensitivity Analysis of Distillation Weights and Robust Consistency

The distillation weight sensitivity experiment shows that as the distillation weight increases, KD Alignment performance improves and peaks at a moderate value before slightly declining, indicating that a balanced weight allows the student model to best inherit the teacher's semantic boundaries without overfitting. For Noise Robustness, performance steadily increases and then stabilizes with higher distillation weights, suggesting that moderate distillation provides consistent supervision, improves reliability under noisy inputs, and enhances overall robustness and adaptability in safety alignment fine-tuning.

The results for Alignment Stability show a generally steady upward trend as the weight increases, with the best value reached within a specific range. This indicates that under distillation constraints, the student model gradually forms more stable alignment outputs. It can maintain consistency under diverse inputs and potential disturbances. This outcome aligns with the core goal of safety alignment, which is to ensure stable and reliable outputs in complex environments without abnormal shifts caused by minor noise or distribution changes.

For the Overall Safety Score, the model shows a trend similar to KD Alignment. The best performance is achieved at a medium weight, followed by a slight decline. This result suggests that overall safety requires a balance between distillation alignment and robustness improvement. An appropriate distillation weight strengthens safety through knowledge transfer while cooperating with noise-robust training to maintain stability. This balance leads to the best comprehensive performance and highlights the effectiveness of the proposed method in combining safety alignment with robust training.

This paper also conducts comparative experiments on the environmental sensitivity of the co-optimization of distillation and robust training based on computational budget and mixed precision settings. The experimental results are shown in Figure 3.

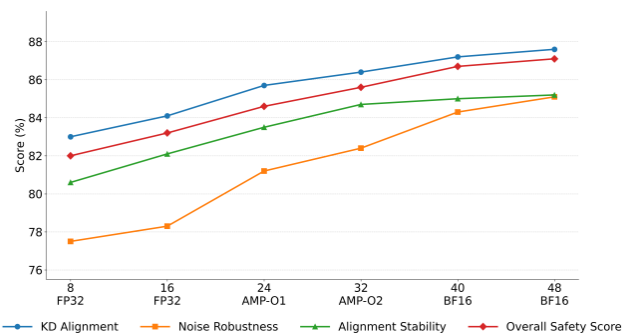

Figure 3. Analysis of the sensitivity of computational budget and mixed precision settings to the co-optimization of distillation and robust training

As the computation budget increases, the KD Alignment metric steadily rises and eventually saturates, indicating that distillation constraints provide alignment gains under limited resources, but plateau with higher budgets, necessitating additional mechanisms for further improvement. Noise Robustness improves even more sharply, especially after introducing mixed precision, showing that robustness training significantly benefits from efficient computation and optimization, resulting in much greater model stability against uncertainty. Alignment Stability also increases gradually, plateauing at high budgets, suggesting that while resource expansion and mixed precision enhance output consistency, gains become marginal beyond a certain point. The Overall Safety Score reflects these patterns, with a marked improvement when switching to mixed precision before stabilizing, demonstrating that the proposed method effectively balances computational cost and alignment robustness, achieving strong safety and efficiency through the integration of distillation and robustness training.

V. CONCLUSION

This study conducts a systematic exploration of safety alignment fine-tuning by comparing distillation and noise-robust training. The proposed framework demonstrates high stability and reliability in both knowledge transfer and adversarial conditions. Through the distillation mechanism, the student model effectively inherits the knowledge boundaries of the teacher model and avoids semantic drift and alignment bias that may occur in conventional training. At the same time, the introduction of noise-robust training allows the model to maintain consistent predictive outputs in complex environments, which significantly enhances overall safety. This method offers a new perspective for fine-tuning large-scale language models and provides a feasible solution to the challenge of balancing alignment and robustness.

Experimental results show that under the combined effects of distillation constraints and robust optimization, the proposed method improves multiple core metrics. The model achieves strong alignment accuracy and maintains stable performance under high noise and uncertain conditions. This ability is essential for models applied in high-risk scenarios, such as privacy-sensitive interactive systems, open environments with frequent adversarial interference, and complex tasks involving multimodal integration. The method effectively mitigates the conflict between safety and adaptability in existing approaches and lays a foundation for building more trustworthy artificial intelligence systems.

At the application level, the outcomes of this study demonstrate significant potential for generalization. The fine-tuning method that balances safety alignment and robustness is not only suitable for natural language processing tasks but can also be extended to medical text analysis, financial risk prediction, and language interfaces in industrial control systems. These domains place strict requirements on the stability and safety of model outputs. The proposed framework can suppress potential risks while maintaining performance. Therefore, this work advances the integration of alignment and robustness research and provides a reference path for real-world cross-domain applications.

Future research can expand in several directions. One line of inquiry is how to extend the combination of distillation and robust training to larger-scale multi-task and multimodal systems while maintaining parameter efficiency. Another is the need to verify safety alignment methods more fully in dynamic environments and adaptive scenarios, where data distributions and adversarial threats change continuously. In addition, combining alignment with interpretability and verifiability research to develop more transparent mechanisms will further improve trust in key societal applications. This study provides theoretical and methodological support for future development and is expected to have a profound impact on broader practical applications.